\documentclass[10pt,twocolumn,letterpaper]{article}

\usepackage{wacv}
\usepackage{times}
\usepackage{epsfig}
\usepackage{graphicx}
\usepackage{amsmath}
\usepackage{amssymb}
\usepackage{booktabs}
\usepackage{subfigure}
\usepackage{cuted}
\usepackage{comment}
\usepackage{color}
\usepackage{mathtools}
\usepackage{caption}
\usepackage{soul}
\usepackage{ulem}
\usepackage{float}

\def\wacvPaperID{1646} 

\wacvalgorithmstrack   

\wacvfinalcopy 

\ifwacvfinal
\usepackage[breaklinks=true,bookmarks=false]{hyperref}
\else
\usepackage[pagebackref=true,breaklinks=true,colorlinks,bookmarks=false]{hyperref}
\fi

\usepackage[capitalize]{cleveref}
\usepackage{multirow}
\usepackage{rotating}

\begin{document}

\newif\ifdraft
\draftfalse

\definecolor{orange}{rgb}{1,0.5,0}
\definecolor{violet}{RGB}{70,0,170}
\definecolor{magenta}{RGB}{170,0,170}
\definecolor{dgreen}{RGB}{0,150,0}
\definecolor{dyellow}{RGB}{150,150,0}
\definecolor{dvio}{RGB}{150,0,150}

\ifdraft
 \newcommand{\PF}[1]{{\color{red}{\bf PF: #1}}}
 \newcommand{\pf}[1]{{\color{red} #1}}
 \newcommand{\WL}[1]{{\color{blue}{\bf WL: #1}}}
 \newcommand{\wl}[1]{{\color{blue} #1}}
  \newcommand{\ME}[1]{{\color{dgreen}{\bf ME: #1}}}
 \newcommand{\me}[1]{{\color{dgreen} #1}}
\newcommand{\HX}[1]{{\color{dyellow}{\bf HX: #1}}}
\newcommand{\hx}[1]{{\color{dyellow} #1}}
\newcommand{\ZY}[1]{{\color{dvio}{\bf ZY: #1}}}
\newcommand{\zy}[1]{{\color{dvio} #1}}
\else
 \newcommand{\PF}[1]{}
 \newcommand{\pf}[1]{#1}
 \newcommand{\WL}[1]{}
 \newcommand{\wl}[1]{#1}
 \newcommand{\ME}[1]{}
  \newcommand{\me}[1]{#1}
  \newcommand{\HX}[1]{}
  \newcommand{\hx}[1]{#1}
  \newcommand{\ZY}[1]{}
  \newcommand{\zy}[1]{#1}
\fi

\newcommand{\parag}[1]{\vspace{-3mm}\paragraph{#1}}
\newcommand{\sparag}[1]{\subparagraph{#1}}
\renewcommand{\floatpagefraction}{.99}

\newcommand{\bA}{\mathbf{A}}
\newcommand{\bC}{\mathbf{C}}
\newcommand{\bD}{\mathbf{D}}
\newcommand{\bI}{\mathbf{I}}
\newcommand{\bH}{\mathbf{H}}
\newcommand{\bP}{\mathbf{P}}
\newcommand{\bR}{\mathbf{R}}
\newcommand{\bZ}{\mathbf{Z}}

\newcommand{\real}{\mathbb{R}}

\newcommand{\bc}{\mathbf{c}}
\newcommand{\f}{\mathbf{f}}
\newcommand{\m}{\mathbf{m}}
\newcommand{\s}{\mathbf{s}}
\newcommand{\bu}{\mathbf{u}}
\newcommand{\x}{\mathbf{x}}
\newcommand{\y}{\mathbf{y}}
\newcommand{\z}{\mathbf{z}}
\newcommand{\w}{\mathbf{w}}

\newcommand{\radius}{\mathbf{r}}

\newcommand{\cF}{\mathcal F}
\newcommand{\fd}{\mathcal{F}_{d}}
\newcommand{\fz}{\mathcal{F}_{z}}

\newcommand{\OURS}[0]{\textbf{OURS}}
\newcommand{\FGSMU}[1]{\textbf{FGSM-U(#1)}}
\newcommand{\FGSMT}[1]{\textbf{FGSM-T(#1)}}
\newcommand{\FGSMUE}[1]{\textbf{FGSM-UE(#1)}}
\newcommand{\FGSMTE}[1]{\textbf{FGSM-TE(#1)}}

\newcommand\extrafootertext[1]{%
    \bgroup
    \renewcommand\thefootnote{\fnsymbol{footnote}}%
    \renewcommand\thempfootnote{\fnsymbol{mpfootnote}}%
    \footnotetext[0]{#1}%
    \egroup
}

\newcommand{\colvecTwo}[2]{\ensuremath{
		\begin{bmatrix}{#1}	\\	{#2}	\end{bmatrix}
}}
\newcommand{\colvec}[3]{\ensuremath{
		\begin{bmatrix}{#1}	\\	{#2}	\\	{#3} \end{bmatrix}
}}
\newcommand{\colvecFour}[4]{\ensuremath{
		\begin{bmatrix}{#1}	\\	{#2}	\\	{#3} \\	{#4}	\end{bmatrix}
}}

\newcommand{\rowvecTwo}[2]{\ensuremath{
		\begin{bmatrix}{#1}	&	{#2}	\end{bmatrix}
}}
\newcommand{\rowvec}[3]{\ensuremath{
		\begin{bmatrix}{#1} &	{#2}	&	{#3} \end{bmatrix}
}}
\newcommand{\rowvecFour}[4]{\ensuremath{
		\begin{bmatrix}{#1}	&	{#2}	&	{#3} &	{#4}	\end{bmatrix}
}}

\newcommand{\tr}{^\intercal}

\newcommand{\rg}{\mathbf{rk}}
\newcommand{\fA}{f_{\mbs \Theta_A}}
\newcommand{\fB}{f_{\mbs \Theta_B}}
\newcommand{\tA}{\mbs \Theta_A}
\newcommand{\tB}{\mbs \Theta_B}

\newcommand{\PPc}[1]{\textcolor{orange}{[\textbf{Pat}: #1]}}
\newcommand{\PP}[1]{\textcolor{orange}{#1}}
\newcommand{\MCc}[1]{\textcolor{blue}{[\textbf{Matt}: #1]}}
\newcommand{\MC}[1]{\textcolor{blue}{#1}}
\newcommand{\MEc}[1]{\textcolor{green}{[\textbf{Martin}: #1]}}
\newcommand{\LC}[1]{\textcolor{cyan}{#1}}
\newcommand{\mbf}[1]{\ensuremath{\mathbf{#1}}}
\newcommand{\mbs}[1]{\ensuremath{\boldsymbol{#1}}}
\newcommand{\mcl}[1]{\ensuremath{\mathcal{#1}}}
\newcommand{\mrm}[1]{\ensuremath{\mathrm{#1}}}
\newcommand{\mbb}[1]{\ensuremath{\mathbb{#1}}}
\newcommand{\msf}[1]{\ensuremath{\mathsf{#1}}}

\newcommand{\ve}[1]{\ensuremath{\mathbf{#1}}} 
\newcommand{\ma}[1]{\ensuremath{\mathsf{#1}}} 

\newcommand{\trsp}[1]{\ensuremath{#1^{\top}}}
\newcommand{\pinv}[1]{\ensuremath{#1^{\dagger}}}
\newcommand{\bmat}[4]{\ensuremath{\begin{bmatrix}#1&#2\\#3&#4\end{bmatrix}}}
\def\trace{\ensuremath{\mathrm{trace}}}
\def\deter{\ensuremath{\mathrm{det}}}
\def\diag{\ensuremath{\mathrm{diag}}}
\def\rank{\ensuremath{\mathrm{rank}}}
\def\Id{\m{Id}} 
\def\mA{\m{A}}
\def\mB{\m{B}}
\def\mC{\m{C}}
\def\mD{\m{D}}
\def\mE{\m{E}}
\def\mF{\m{F}}
\def\mG{\m{G}}
\def\mH{\m{H}}
\def\mK{\m{K}}
\def\mL{\m{L}}
\def\mN{\m{M}}
\def\mP{\m{P}}
\def\mW{\m{W}}
\def\mX{\m{X}}
\def\mY{\m{Y}}
\def\mZ{\m{Z}}

\newcommand{\bvec}[2]{\ensuremath{\begin{bmatrix}#1\\#2\end{bmatrix}}}
\def\One{\mbs{1}} 
\def\va{\ve{a}}
\def\vb{\ve{b}}
\def\vc{\ve{c}}
\def\vd{\ve{d}}
\def\vf{\ve{f}}
\def\vg{\ve{g}}
\def\vh{\ve{h}}
\def\vi{\ve{i}}
\def\vt{\ve{t}}
\def\bx{\ve{x}}
\def\by{\ve{y}}
\def\bz{\ve{z}}
\def\bv{\ve{v}}

\def\cL{\mcl{L}}

\def\ie{\emph{i.e.}}
\def\eg{\emph{e.g.}}
\def\iid{\emph{i.i.d.}}
\def\wrt{w.r.t.}
\def\mwrt{\mrm{w.r.t.}}
\def\msbt{\mrm{sb.t.}}

\def\sqt{^{\frac{1}{2}}} 
\def\msqt{^{-\frac{1}{2}}} 
\def\R{\mbb R}
\def\vtheta{\mbs{\theta}}

\title{Two-level Data Augmentation for Calibrated Multi-view Detection}

\author{Martin Engilberge\thefootnote{*} \hspace{1.4em} Haixin Shi\thefootnote{*} \hspace{1.4em} Zhiye Wang \hspace{1.4em} Pascal Fua\\
EPFL, Lausanne, Switzerland \\
{\tt\small firstname.lastname@epfl.ch} 
}

\maketitle
\thispagestyle{empty}

\begin{abstract}
   
Data augmentation has proven its usefulness to improve model generalization and performance. While it is commonly applied in computer vision application when it comes to multi-view systems, it is rarely used. Indeed geometric data augmentation can break the alignment among views.
This is problematic since multi-view data tend to be scarce and it is expensive to annotate.

In this work we propose to solve this issue by introducing a new multi-view data augmentation pipeline that preserves alignment among views. Additionally to traditional augmentation of the input image we also propose a second level of augmentation applied directly at the scene level. 
When combined with our simple multi-view detection model, our two-level augmentation pipeline outperforms all existing baselines by a significant margin on the two main multi-view multi-person detection datasets WILDTRACK and MultiviewX.

\end{abstract}

\def\thefootnote{*}\footnotetext{These authors contributed equally to this work.}\def\thefootnote{\arabic{footnote}}
\extrafootertext{Project code at \url{https://github.com/cvlab-epfl/MVAug}}

 \section{Introduction}\label{sec:intro}
  
\normalem

In recent years deep learning models have been widely adopted in the computer vision fields. One of the reasons for this wide adoption is the generalization ability of gradient-based models \cite{kawaguchi2017}. While such models generalize well, they are still subject to overfitting their training data. Multiple methods have been proposed to combat overfitting. Some focus on the model design \eg~dropout layer \cite{Srivastava14} or batch normalization \cite{Ioffe15}, while others such as data augmentation \cite{shorten2019survey} directly tackle one of the root causes of overfitting: overparametrization due to limited data.
While data augmentation has been widely used and studied in a variety of fields, it is rarely used in the multi-view context. Indeed in a multi-view setup geometric data augmentation can easily break the alignment among views.


  
 \begin{figure}
  \centering
  \includegraphics[width=\linewidth]{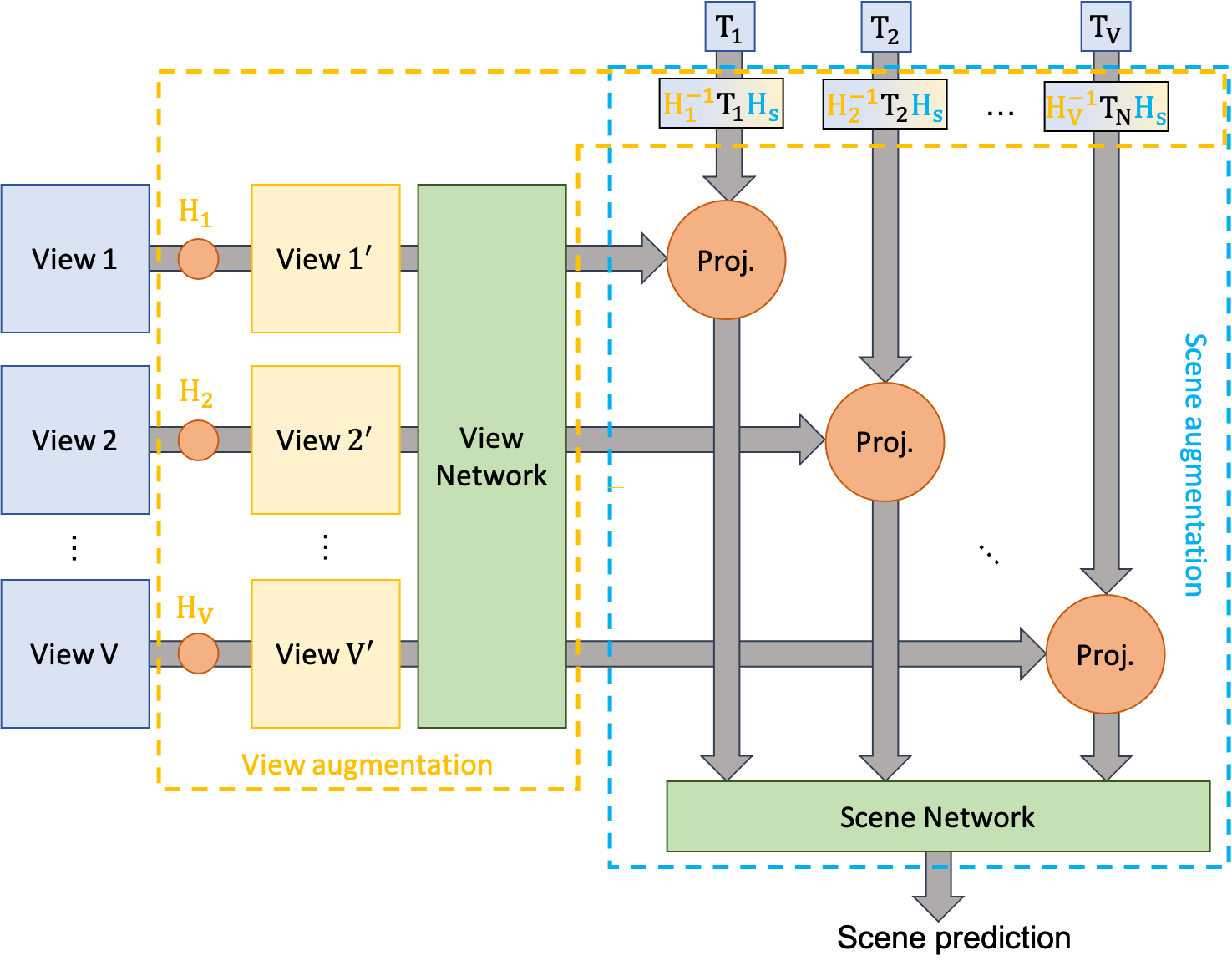}
  \caption{\textbf{Data augmentation in a multi-view setting} Illustration of a multi-view model combined with our multi-view data augmentation pipeline. The input of the model consists of multiple images coming from different viewpoints.
  Each view is associated with a transformation $\ma T_v$ that projects the corresponding view to a common scene representation where the different views are aligned.
  In yellow, our view based augmentation mechanism applies data augmentation ($\ma H_{v}$) independently on each view and updates the original transformation to preserve alignment. It helps reduce the view network from overfitting the training data.
  Circled in light blue, our new scene augmentation is applied directly in the aligned scene representation by updating the projection transformation $\ma T_v$ with a scene augmentation $\ma H_S$.  
  }
  \label{figure:intro}
\end{figure}

Most multi-view people detection methods do not employ data augmentation \cite{Xu16, Fleuret08a,Chavdarova17}. While less than ideal, this wasn't the main limiting factor in those earlier methods, since they only used deep learning models for initial monocular predictions which could be pre-trained using monocular data augmentation. However recent approaches \cite{song2021stacked, hou2020multiview} have adopted end-to-end architectures directly predicting detections in the ground plane (top view) from multi-view inputs. When trained from scratch, such methods can greatly benefit from data augmentation.

In this paper we propose to address the issue by introducing a data augmentation pipeline for multi-view model illustrated in \cref{figure:intro}. Our pipeline is able to augment each view independently while preserving the overall alignment among views (view augmentation). Additionally we introduce a new type of multi-view augmentation, applied directly at the scene level, we call it scene augmentation. Each type of augmentation helps reduce overfitting of different parts of the network. We demonstrate the benefit of both types of augmentation on the multi-view multi-object detection task and show that when combined with our model it outperforms state-of-the-art multi-view methods \cite{hou2020multiview,song2021stacked,hou2021multiview} on the challenging WILDTRACK \cite{Chavdarova18a} and MultivievX \cite{hou2020multiview} datasets.

\ULforem

  
 \begin{figure}
  \centering
  \includegraphics[width=\linewidth]{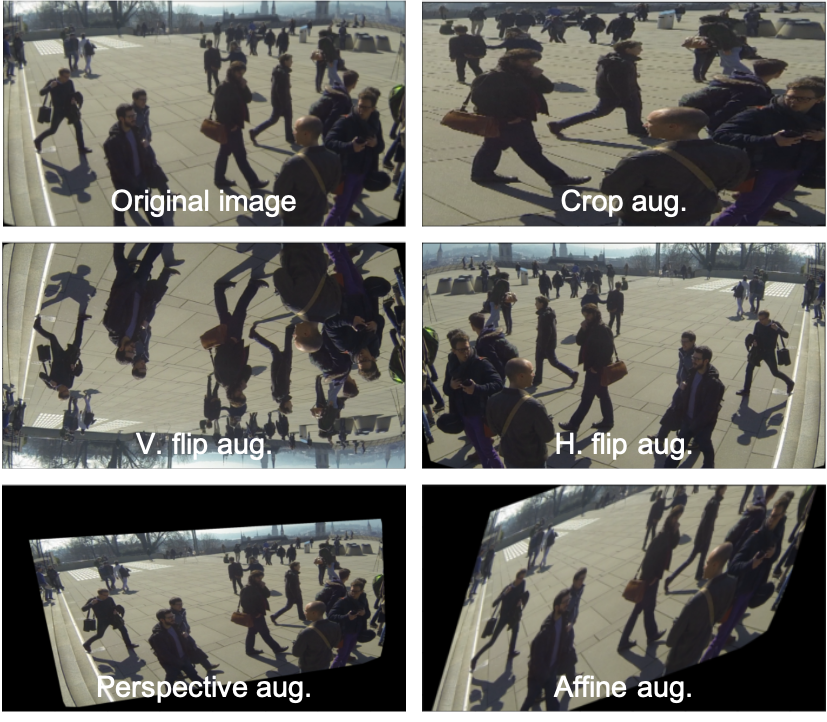}
  \caption{\textbf{Geometric data augmentation} Visualization of the different kinds of geometric data augmentations, top left is the un-augmented image.}
  \label{figure:visu_aug_type}
\end{figure}

 \section{Related works}\label{sec:related}
 In this section we briefly introduce previous work on multi-view detection and discuss existing approaches for data augmentation.

\vspace{0.5cm}
\textbf{Multi-view multi-person detection} Over the year multiple monocular detection methods have been proposed such as the R-CNN family of models \cite{Girshick13, Ren15} that predict bounding boxes from single input image using a two-stage architecture. More recently single stage anchor-free approaches \cite{Zhou19,Tian19b} have yielded promising results. 

However when it comes to detecting people in crowded scene they tend to miss heavily occluded people. To remedy this problem multiple works have proposed to detect people in a multi-view setup. Using multiple calibrated cameras \cite{Tsai87} reduces the likelihood to suffer from occlusion in every view.
To aggregate multiple views most existing methods predict pedestrian occupancy map on the ground plane \cite{Fleuret08a, Baque17b, Chavdarova17, hou2020multiview, hou2021multiview, song2021stacked}. While final detections are done on the ground plane, some models first predict monocular detections \cite{Xu16,Fleuret08a} before projecting and aggregating the results. Others choose to combine view aggregation and prediction in a single step by for example learning jointly a CNN and Conditional Random Fields (CRF) \cite{Baque17b,Roig11}. More recent approaches learn end-to-end neural networks, where projections on the ground plane are part of the networks. One such approach \cite{hou2021multiview} proposes a view aggregation network that leverage an attention mechanism as part of a transformer network to select the most relevant part of each view to generate the final detection map. In \cite{song2021stacked} instead of a single projection on the ground plane, they propose to use multiple projections onto planes at different heights in order to approximate a 3d world coordinate system.

\vspace{0.5cm}
\textbf{Data Augmentation} Data augmentation is widely used to improve generalization of neural networks \cite{shorten2019survey}.
During training it provides artificial samples generated by altering original data in multiple ways. Traditional methods can be roughly divided into two sorts. First, the geometric transformations methods, including flipping, cropping, rotation and translation, tackle positional bias in training data. The other is photometric transformations which performs augmentations in the color channels space or injects noise into images\cite{DBLP:journals/jbd/ShortenK19}. With the booming of deep learning, many image data augmentation methods combined with deep learning have been developed. Feature space augmentation proposed by DeVries and Taylor \cite{DBLP:conf/iclr/DeVriesT17} extracts vectors from low-dimensional feature maps and adds noise, interpolates, and extrapolates. Adversarial training uses the samples generated by the rival network for augmentation \cite{DBLP:journals/jbd/ShortenK19}.

Image augmentation in detection setting faces multiple challenges. When bounding boxes ground truth are used, data augmentation cannot be applied directly, special augmentation is required to correctly preserve ground truth boxes \cite{DBLP:conf/eccv/ZophCGLSL20}.
\normalem
Anchor-free detection models are free of such limitation, however, when used in a multi-view setting combining them with data augmentation can be the cause of inconsistency among views. For such reasons traditional geometric image augmentations are rarely used in multi-view settings\cite{DBLP:conf/3dim/WangS18,hou2020multiview}. To ensure the alignment among different views in multi-view pedestrian detection, Hou \emph{et~al.} \cite{hou2021multiview} proposed to augment each view individually via geometric transformation and then reverse the augmentation before projection.
\ULforem Having to reverse the data augmentation is one drawback of such approach, by doubling the number of projection happening in the network it introduces noises in the features due to repeated bilinear interpolation.

\begin{figure*}
    \centering
    \includegraphics[width=\linewidth]{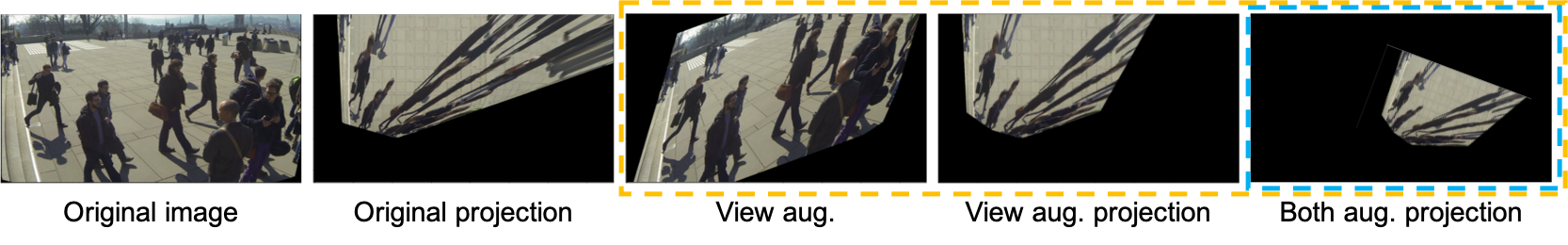}
    \caption{\textbf{Visualization of view and scene augmentation, and their effect on the ground plane projection} The two left images correspond to the original image and its corresponding projection on the ground plane. The third and fourth images visualize the effect of affine view augmentation, and the projection on the ground plane of the augmented image. Note that between second and forth image the alignment is preserved. The last image visualizes the effect of adding affine scene augmentation to the view augmented image on the ground plane projection.
    }
    \label{figure:model_visu_aug}
  \end{figure*}

 \section{Approach}\label{sec:approach}
 We tackle the multi-view multi-person detection problem. In this section we introduce the problem formalism.
Then we present our multi-view data augmentation framework. Finally, we show how it is combined with our multi-view network.

\subsection{Multiview detection formalism}
Let us consider a scene containing $V$ different cameras with partially overlapping fields of view.
Each camera is calibrated \cite{Tsai87}, yielding the calibration $\mathbf{C_v} = \left\{\ma K_v, \ma R_v, \ve t_v \right\}$. Where $\ma K_v$ is the intrinsic camera matrix, and $\ma R_v$ and $\ve t_v$ are the extrinsic camera parameters.

A set of frames $\mathbf{I}=\left\{\mathbf{I}_1, \ldots \mathbf{I}_V\right\}$ coming from the different cameras can be projected to a common ground plane using a top view reprojection. The top view projection matrix $\ma T_v$ for view $v$ can be derived from the calibration as follows $\ma T_v = \ma K_v[\ma R_v |\ve t_v ]$ assuming that the ground plane has a zero z-coordinate ($z=0$) in the world coordinate system.
The projection of an image onto the ground plane is then written as $\mathbf{I}^{ground}_v = P(\mathbf{I}_v, \ma T_v )$ where $P$ is the projection function.

\subsection{Geometric data augmentation}

Applying data augmentation in a multi-view context is not trivial, when applying geometric transformation to an image it invalidates its calibration and with it the projection on the common ground plane. We propose to solve this issue by extending the augmentation process to include the ground plane projection matrix $\ma T_v$.

We focus our attention on the following geometric data augmentation: flipping, cropping, affine transformation and perspective transformation. An example of each transform is visible in \cref{figure:visu_aug_type}. It is possible to express all these geometric data augmentation in the form of a homography $\ma H$. The Appendix Section 1 contains the detailed homographies for each augmentation.

\paragraph{View augmentation} Our approach consists of two types of augmentation, first the one we call view augmentation, which is applied to the input image. This is similar to standard data augmentation. However we also update the ground plane projection in order to preserve the alignment among views. Note that each view is augmented independently and can be transformed by different augmentations.

Given a homography $\ma H_v$ characterizing a view augmentation and the image $\mathbf{I}_v$ for view $v$ we write the augmented image as $\mathbf{I'}_v = P(\mathbf{I}_v, \ma H_v$). The augmented ground plane projection is written as 
$\ma T_v' = \ma H^{-1}_v \ma T_v$. In a single step the augmented projection, inverse the effect of the image data augmentation before doing the original ground plane projection.

\begin{figure*}[t]
  \centering
  \includegraphics[width=\linewidth]{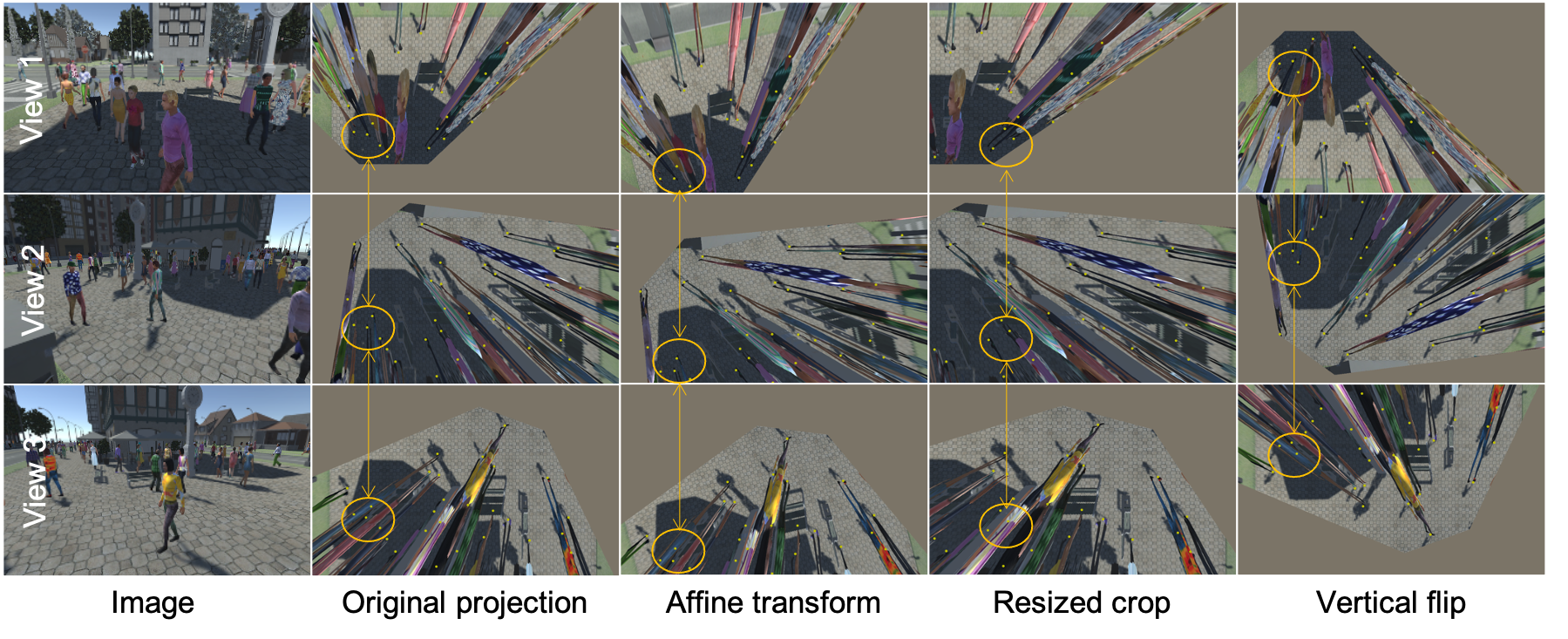}
  \caption{\textbf{Visualization of scene augmentation} We visualize the effect of different scene augmentation by projecting the original image onto the ground plane using ground plane homography augmented with scene augmentation. Note that the ground plane augmentation is the same for all the views which guarantees to preserve the alignment between views. The orange circle highlight the same ground truth points across the two views.}
  \label{figure:visu_scene_aug}
\end{figure*}

\paragraph{Scene augmentation} The second type of augmentation is novel and specific to multi-view training, we call it scene augmentation. Scene augmentation only changes the ground plane projection matrix, it modifies the projection of all the views in a similar manner. Intuitively it can be seen as applying augmentation on the ground plane directly. In practice it consists only in a modification to the ground plane projection, and doesn't require any additional projection step. We visualize the effect of scene augmentation in \cref{figure:visu_scene_aug}.


  
Given a scene augmentation characterized by the homography $\ma H_S$ the ground plane projection is augmented as follows $\ma T_v' = \ma T_v \ma H_S $. Note that scene augmentation is independent of the viewpoint, all views are augmented with the same $\ma H_S$.

Both types of augmentation modify the ground plane projection matrix but they do it independently, it is possible to apply each type of augmentation on their own, or to combine the two. When both are applied, the augmented ground plane projection can be written as $\ma T_v' = \ma H^{-1}_v \ma T_v \ma H_S $. \cref{figure:model_visu_aug} contains a visualization of both types of augmentation.

\subsection{Model Architecture and Training}

\begin{figure*}
    \centering
    \includegraphics[width=\linewidth]{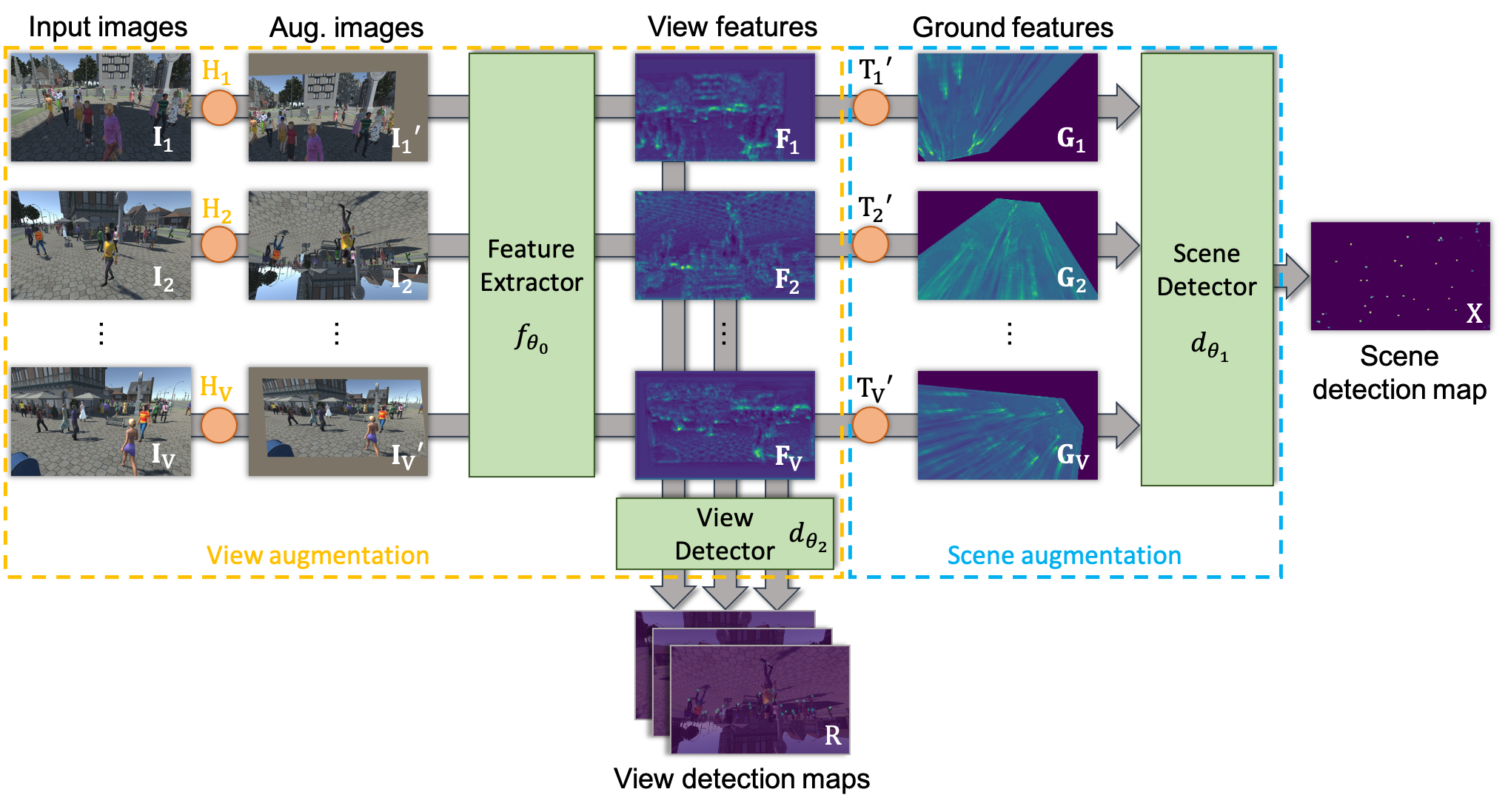}
    \caption{\textbf{Details of the proposed multi-view multi-person detection architecture} A set of input images $\mathbf{I}$ is augmented with the proposed view augmentation. Each view augmentation is reflected on the set of ground plane projection homographies $\mathbf{T}$ to form the augmented homographies $\mathbf{T'}$ preserving the alignment on the ground plane. The augmented images go through a feature extractor module, then the features are being projected onto the ground plane where they are aggregated by the scene detector which outputs the final scene detection heatmap. In parallel, features from individual images are fed to the view detector to generate view detection used for regularization purposes. Additionally ground plane projection homographies $\mathbf{T}$ can be extended with our second type of augmentation, scene augmentation, which applies augmentation directly in the ground plane.
    Boxes with green background correspond to learnable modules, arrows going through a module represent elements being processed independently by that module. Orange discs represent projection operation, the letter above each disc corresponds to the homography used by the projection.}
    \label{figure:model}
  \end{figure*}

In this section we describe the architecture and training procedure of our multi-view detection model. The overall architecture can be seen in \cref{figure:model}

\paragraph{Multi-view detection}

The proposed multi-view model consists of three learnable modules, first the feature extractor based on a truncated ResNet 34 \cite{He16a} process each image independently. Each feature is then projected on the ground plane using its associated projection matrix. After projection the features are concatenated and goes through the scene detector which outputs the final scene detection map.
More formally, it reads as follows:

\begin{equation}
  \ve I' \xmapsto{f_{\mbs\theta_0}} 
  \ve F \xmapsto{P(\ve F, \ve T' )} 
  \ve G \xmapsto{d^S_{\mbs\theta_1}}
  X
  \label{eq:pipe_scene}
\end{equation}

Where $\ve I'$ is the set of augmented input images and $\ve T'$ is its corresponding set of augmented top view projection matrices.
$ \ve F = \left\{\mathbf{F}_1, \ldots \mathbf{F}_V\right\} $ is the output of the ResNet parameterized by weights $\theta_0$, with view features $\ve F_v = f_{\mbs\theta_0}( \ve I_v')$. $\ve G$ corresponds to the projected features in the ground plane $ \ve G = \left\{\mathbf{G}_1, \ldots \mathbf{G}_V\right\}$ with $\ve G_v = P(\ve F_v, \ma T_v')$. 
 Finally, the ground features are concatenated and goes through the scene detector parameterized by weights $\theta_1$ which outputs the final detection heatmap on the ground plane. We denote $X = F(\ve I', \ve T', \theta_0, \theta_1)$ for short this scene detection pipeline.

In parallel, the image features $\mathbf{F}$ of all the views go into the view detector which processes them independently and outputs a view detection map for each of them.
It reads as follows:
  \begin{equation}
    \ve F_v \xmapsto{d^V_{\mbs\theta_2}}
    R_v
    \label{eq:pipe_view}
  \end{equation}
Where $\mathbf{F}_v$ is the ResNet output defined above. $R_v$ corresponds to the detection heatmap in the image plane for view $v$. It is the output of the view detector parameterized by weights $\theta_2$. We denote $R_v = F(\ve I', \theta_0, \theta_2)$ for short this view detection pipeline.

\paragraph{Loss function}
The aim of the training is to learn the weights $\theta_{0:2}$ of the model. Given the model inputs $\ve I'$ and $\ve T'$  and their corresponding scene detection ground truth $\hat{X}$ and view detection ground truth $\hat{R}_v$ our model is trained with two loss functions.

For the scene detection loss, we use the  Mean Squared Error (MSE) defined as follows:
\begin{equation}
  \mathcal{L}_{\text {ground}}(X, \hat{X}) = \left( X - \hat{X} \right)^{2} 
  \label{eq:lossdetground}
\end{equation}
Similarly to \cite{hou2020multiview} we also supervised the training directly in the image plane with the following loss.
\begin{equation}
  \mathcal{L}_{\text {image}}(R, \hat{R}) = \frac{1}{V}  \sum_{v=1}^{V} \left(R_v - \hat{R}_v\right)^{2}	 
  \label{eq:lossdetimage}
\end{equation}
As opposed to \cite{hou2020multiview} we only apply this loss for detection at feet level instead of both feet and head. We found empirically no benefit for adding additional supervision at head level. $\mathcal{L}_{\text {image}}$ serves two purposes, first it acts as a regularizer pushing the feature extractor to generate relevant features independently for each view. Secondly, when combined with view augmentation, it helps reduce overfiting in the feature extractor part of the model.

Both losses are summed to form the training loss $$ \mathcal{L} = \mathcal{L}_{\text {ground}} + \mathcal{L}_{\text {image}}.$$

 \section{Experiments}\label{sec:expe}
 We validate our approach on the multi-view multi-person detection task using the WILDTRACK and MultiviewX datasets.

\begin{table*}[h]
    \begin{center}
    \renewcommand{\arraystretch}{1.3}
    \begin{tabular}{r  c c c c  c c c c c} \toprule
    &  \multicolumn{4}{c}{WILDTRACK dataset} & & \multicolumn{4}{c}{MultivievX dataset} \\ \cmidrule(lr{.75em}){2-5} \cmidrule(lr{.75em}){7-10}
    model & MODA & MODP & Prec. & Rec. & & MODA & MODP & Prec. & Rec.  \\ \midrule
    DeepOcclusion \cite{Chavdarova18a} & 74.1 & - & 95.0 & 80.0 & & - & - & -  \\
    MVDet \cite{hou2020multiview} & 88.2 & 75.7 & 94.7 & 93.6 & & 83.9 & 79.6 & 96.8 & 86.7 \\
    SHOT \cite{song2021stacked} & 90.2 & 76.5 & 96.1 & 94.0  & & 88.3 & 82.0 & 96.6 & 91.5 \\
    MVDeTr \cite{hou2021multiview} & 91.5 & \textbf{82.1} & \textbf{97.4} & 94.0 & & 93.7 & \textbf{91.3} & \textbf{99.5} & 94.2  \\
    MVAug (Ours) & \textbf{93.2} & 79.8 & 96.3 & \textbf{97.0} & & \textbf{95.3} & 89.7 & 99.4 & \textbf{95.9} \\
    \bottomrule
    \end{tabular}
    \end{center}
      \caption{\textbf{Multi-view multi-person detection}  Detection performance of our proposed model on the WILDTRACK and MultiviewX datasets. We report MODA, MODP, precision and recall \cite{kasturi09}. The proposed approach outperforms all existing baseline in terms of MODA on both datasets. In general this performance gain can be explained by a increase in recall.
      }  
     \label{tab:sotadetection}
    \end{table*}
\subsection{Experimental setup}

\paragraph{Datasets} To train our model we use two multi-view pedestrian datasets: The WILDTRACK dataset has 7 cameras that focus on an area of $12\text{m} \times 36\text{m}$ in the real world. It contains 400 synchronized frames per view with a resolution of $1080 \times 1920$. Each person is annotated with a bounding box. \cref{figure:model_visu_aug} shows an image from the WILDTRACK dataset.

The MultiviewX dataset has 6 cameras that focus on an area of $16\text{m} \times 25\text{m}$. It is a synthetic dataset representing a virtual world. It also contains 400 synchronized frames per view with a resolution of $1080 \times 1920$. For both dataset the images are resized to $536 \times 960$ before being augmented and fed to the model.
Three images coming from different views can be seen in \cref{figure:visu_scene_aug}.

The aggregation of multiple views is done in the ground plane. In WILDTRACK we discretize the ground plane such that one cell correspond to 20 cm resulting in ground plane map of dimensionality $180 \times 80$. For MultiviewX the ground plane map has a dimensionality of $160 \times 250$ with cell corresponding of 10 cm. The scales of the ground plane map have been chosen to minimize computational cost.

\paragraph{Evaluation Metrics} We adopt similar evaluation metrics as previous work \cite{Chavdarova18a,hou2020multiview,hou2021multiview}, we report Precision, Recall, MODA, and MODP \cite{kasturi09}. A threshold equivalent to 0.5 meters is used to determine true positives. We use the matlab MOTChallenge Evaluation toolkit.

\paragraph{Implementation details}
Our model is implemented in Pytorch, and runs on a single Nvidia v100 GPU. The data augmentation pipeline wraps the original Torchvision augmentation in order to extract their parameters and generate the corresponding homography. During training random affine transformations are used for both view and scene augmentation and in both cases, a proportion of 50\% of the training data is augmented. 

The feature extractor is based on a ResNet 34 with its last four layers removed. It outputs feature of dimensionality 128. 

The view detector consists of two pairs of ReLu and a $1\times1$ convolutional layer followed by a sigmoid function. First convolution layer contains 128 filters and the second one a single filter. The output of the view detector is only used for regularization purposes, hence the minimal architecture of the view detector allows for greater regularization effect on the feature extractor.

For the scene detector, we adopt a multi-scale architecture, this detector is responsible for aggregating the ground plane features coming from multiple views. Therefore it needs to be able to handle slight misalignment among them due to calibration error. The scene detector consists of four scales where the spatial resolution of the features is halved in among each scale using adaptive average pooling. Each  scale consists of four blocks of convolutional layer - batch normalization \cite{Ioffe15} - ReLu \cite{Nair10}. The output of the four scales are bilinearly interpolated back to their original dimension, concatenated and fed into a final $1\times1$ convolutional layer followed by a sigmoid function to produce the final scene detection heatmap.

Our model outputs probabilistic detection heatmaps, to compute evaluation metrics we extract detection points from those heatmaps. We apply Non Maximum Suppression (NMS), then select the top 200 detections and use K-Means clustering on detection score with K=2 to separate true detection from noise.

\paragraph{Augmentation parameters}
We list the parameters used for each type of geometric transforms. For random affine augmentation, the rotation can be up to 45 degrees, the translation up to 20\% on both directions, the scaling up to 20\% up or down, and the shearing up to 10 degrees. For the random resized crop, the crop covers an area of 80\% to 100\% of the original image with an aspect ration between 0.75 and 1.33 before being resized to the original image size. The perspective transformation uses a distortion scale of 0.5. Horizontal and vertical flips don't require any parameters.

\subsection{Comparing to the State-of-the-Art}

On the multi-view people detection task, we compare our model to 4 baselines. Results can be found in \cref{tab:sotadetection}. On both WILDTRACK and MultiviewX the proposed model using our two-level augmentation scheme outperforms all previous baseline on MODA with a significant margin. In particular it outperforms MVDeTr which uses a simpler form of view based augmentation combined with a more complex transformer based architecture. The improvement in MODA can be explained by an increase in recall, in general our model detects people that were missed by other models. It confirms the better generalization of our model due to our data augmentation pipeline.

Note that our model underperforms on the MODP metric when compared to MVDeTr this can be explained by our choice of ground plane discretization strategy. MVDeTr uses much smaller cells of 2.5 cm. Even though the metric threshold has been adjusted to account for this, rounding error from the change of scale remains and mostly affect MODP which is directly computed from distances in the discretized space.
As stated above the coarser grid was chosen for computational reason due to the large number of experiments needed to evaluate the proposed data augmentation pipeline.

\vspace{1cm}
\subsection{Further Analysis}
We conduct additional experiments to justify the design choice of our method, and we evaluate the contribution of each of its components. 
In an effort to stay as close as possible to a real-life scenario, all the following experiments are conducted on the WILDTRACK dataset.

\paragraph{Optimal combination of view and scene augmentation}
We propose to investigate which combination of view and scene augmentation is optimal. In \cref{tab:abla_combination} we report the MODA metric for multi-view people detection on WILDTRACK. When only scene augmentation is used, the affine augmentation is most beneficial.
When only view augmentation is used, affine augmentation and crop augmentation perform very well. We can see that when only one type of augmentation is used, view augmentation generates greater improvement than scene augmentation. Overall when compared to no augmentation at all, most augmentation strategies are beneficial. Finally, the best pairwise combination of augmentation consists of using random affine for both view and scene augmentation.

\begin{table}[h]
\begin{center}
\resizebox{\linewidth}{!}{
\renewcommand{\arraystretch}{1.5}
\begin{tabular}{c r | c c c c c c} \toprule
\multicolumn{2}{c}{} &  \multicolumn{6}{c}{View augmentation} \\ 
& & \rotatebox{90}{No aug}  & \rotatebox{90}{H-Flip} & \rotatebox{90}{V-Flip} & \rotatebox{90}{Affine} & \rotatebox{90}{Persp.} & \rotatebox{90}{Crop} \\ \cmidrule{2-8}
\multirow[c]{6}{0cm}{\rotatebox{90}{Scene augmentation}} & No aug & 90.86 & 91.28 & 91.49 & 92.65 & 92.23 & 92.45 \\ 
& H-Flip & 91.39 & 90.65 & 91.81 & 92.44 & 91.70 & 92.54 \\ 
& V-Flip & 90.55 & 91.60 & 90.97 & 91.91 & 91.07 & 92.02 \\ 
& Affine & 91.49 & 91.91 & 92.02 & \textbf{93.17} & 91.49 & 92.44 \\ 
& Persp. & 91.28 & 90.86 & 90.44 & 90.86 & 90.44 & 91.49 \\ 
& Crop & 90.76 & 92.54 & 91.91 & 91.49 & 91.81 & 92.44 \\ \bottomrule
\end{tabular}
}
\end{center}
\caption{\textbf{Combination of view and scene augmentation} We report MODA metric on the WILDTRACK dataset for all pairwise combination of view and scene augmentations. For view augmentation crop, affine, and perspective augmentation perform best. For scene augmentation affine and horizontal flip augmentation are best. Best results is obtained by combining affine view augmentation with affine scene augmentation.
}  
\label{tab:abla_combination}
\end{table}

\paragraph{Ablation study}
We conduct an ablation study to measure how each of the loss $\mathcal{L}_{\text{image}}$, the view augmentation and the scene augmentation contribute to the overall performance of the system. We report MODA and MODP on the WILDTRACK dataset in \cref{tab:abla_wild}. On their own each component improves MODA, both augmentation generate greater improvement than $\mathcal{L}_{\text{image}}$. When combined with view augmentation, $\mathcal{L}_{\text{image}}$ improves MODA by almost a point, whereas it has detrimental effect when combined with scene augmentation alone. When scene augmentation is used, it systematically improves MODP. The best result is obtained when everything is combined.

\begin{table}[h]
    \begin{center}
    \renewcommand{\arraystretch}{1.5}
    \begin{tabular}{c c c | c c } \toprule
    \multicolumn{3}{c}{Model component} &  \multicolumn{2}{c}{Metrics} \\ 
    $\mathcal{L}_{\text {image}}$ & View aug. & Scene aug.  & MODA & MODP\\ \midrule
    &  &  & 90.65 & 76.92\\
    \checkmark & & & 90.86 & 79.59 \\
    &  \checkmark &  & 91.28 & 77.52 \\
    & & \checkmark & 91.28 & 79.20 \\
    \checkmark & \checkmark &  & 92.12 & 77.67 \\
    \checkmark & & \checkmark & 90.65 & 78.41 \\
    \checkmark & \checkmark & \checkmark & \textbf{93.17} & \textbf{79.83} \\

    \bottomrule
    \end{tabular}
    \end{center}
    \caption{\textbf{Ablation results on WILDTRACK} We report MODA and MODP metric on the WILDTRACK dataset, we evaluate the contribution of the proposed view augmentation, scene augmentation and the image prediction loss. Without $\mathcal{L}_{\text {image}}$ loss, both view and scene augmentation perform similarly. When $\mathcal{L}_{\text {image}}$ is added view augmentation perform significantly better.
    The best result is obtained with the combination of the three components.
    } 
    \label{tab:abla_wild}
    \end{table}

\paragraph{Augmentation proportion}
We propose to evaluate how the proportion of augmentation impacts detection results. To do so we vary the percentage of training data that is subjected to either view or scene augmentation. We report MODA and MODP on the WILDTRACK dataset in \cref{tab:prop_wild}. 
When percentage of augmentation is kept identical for both view and scene augmentation, the best result is obtained when 50\% of the training data is augmented. We also test the effect of having different percentages for view and scene augmentation, better detection results are obtained when the proportion of view augmentation is larger than the proportion of scene augmentation (See line two and three of \cref{tab:prop_wild}).

\begin{table}[h]
    \begin{center}
    \renewcommand{\arraystretch}{1.5}
    \begin{tabular}{c c | c c } \toprule
    \multicolumn{2}{c}{} &  \multicolumn{2}{c}{Metrics} \\ 
    View aug. & Scene aug.  & MODA & MODP\\ \midrule
    0\% & 0\% & 90.86 & 79.59\\
    25\% & 25\% & 91.49 & 78.24\\
    25\% & 50\% & 90.86 & 79.41\\
    50\% & 25\% & 92.12 & 79.66\\
    50\% & 50\% & \textbf{93.17} & \textbf{79.83}\\
    75\% & 75\% & 90.65 & 78.47\\
    100\% & 100\% & 90.44 & 78.24\\
    \bottomrule
    \end{tabular}
    \end{center}
    \caption{\textbf{Varying proportion of augmentation} We report MODA metric on the WILDTRACK dataset. We evaluate the effect of varying the proportions of input images affected either by view or scene augmentation. Best result is obtained with 50\% of the training image augmented with both view and scene augmentation. When using unbalanced proportion, it is beneficial to have of view augmentation rate higher than the scene augmentation rate. 
    } 
    \label{tab:prop_wild}
    \end{table}

\paragraph{Effect on overfitting}
The main goal of data augmentation is to increase model generalization by reducing overfitting to the training dataset.
We propose to measure how each component of our approach  helps reduce overfitting. On the WILDTRACK dataset, we measure overfitting by computing the ratio of validation loss over training loss. Ideally, the overfitting ratio should be one, meaning that the model performs similarly on the training and validation dataset.

\begin{figure}
    \centering
    \includegraphics[width=\linewidth]{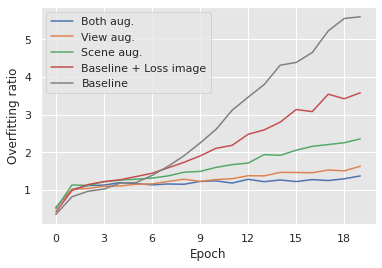}
    \vspace{-1em}
    \caption{\textbf{Effect of data augmentation on training overfitting} We visualize the evolution of the overfitting ratio over training epochs. It is computed by dividing the validation loss by the training loss. Each component of our method contributes in reducing overfitting, the best result is obtained when both augmentations are combined with the image loss.}
    \label{figure:overfitting}
  \end{figure}

We can see in \cref{figure:overfitting} that for our baseline model which uses neither $\mathcal{L}_{\text{image}}$ nor any kind of augmentation, it is not the case and the overfitting ratio quickly grow over 5.
Adding $\mathcal{L}_{\text{image}}$ to the baseline help reduce the ratio, additionally using scene augmentation or view augmentation further reduces overfitting. Note that view augmentation has more impact on overfitting than scene augmentation. Finally, when the three components are used together, the overfitting ratio is reduced the most and very close to an ideal ratio of one.

\section{Limitations}
\paragraph{}

In our experiment we limited ourselves to using only one type of geometric augmentation for both scene and view augmentation. It might be possible to further improve performances by combining multiple type of augmentation. However with more than one type of augmentation the number of possible combinations becomes quite large and therefore computationally expensive to evaluate systematically.

\paragraph{}
Similarly due to limited computational resources and large number of experiments we were only able to run each training once, ideally we would like to average the results over multiple runs. Nonetheless, with the current results we were able to observe general trends when it comes to data augmentation in a multi-view detection system.

 \section{Conclusion}\label{sec:conclusion}
  In this paper, we proposed a new two-level augmentation pipeline for multi-view multi-person detection. When combined with our simple multi-view end-to-end trainable model, it outperforms all existing baselines.
 
Through extensive ablation studies, we show the contribution of each component of our model and their interaction with each other.  We systematically evaluate all pairwise combination of scene and view augmentation. Furthermore we confirm that the proposed approach is effective on real data, by obtaining state-of-the-art results on both the WILDTRACK and MultiviewX datasets.

 \paragraph{Acknowledgments} This work was funded in part by the Swiss Innovation Agency.

\clearpage
 \normalem
{\small
\bibliographystyle{ieee_fullname}
\bibliography{bib/wacv,bib/optim,bib/string,bib/vision,bib/learning}
}

\end{document}